\documentclass[sigconf]{acmart}
\makeatletter
\renewcommand\@formatdoi[1]{\ignorespaces}
\makeatother
\makeatletter
\def\@copyrightspace{
    \@float{copyrightbox}[b]
    \begin{center}
        \setlength{\unitlength}{0.2pc}
            \begin{picture}(100,6) %
                \put(0,-0.95){\crnotice{\@toappear}}
            \end{picture}
        \end{center}
    \end@float}
\makeatother

\usepackage{booktabs} %
\usepackage{graphicx}
\usepackage{pgfplots}
\pgfplotsset{compat=1.13}
\graphicspath{{figs/}}

\usepackage{amsmath,amsfonts,bm,xspace}

\newcommand{\model}{NOCD\xspace}
\newcommand{\modelX}{NOCD-X\xspace}
\newcommand{\modelA}{NOCD-G\xspace}
\newcommand{\deepwalk}{DeepWalk\xspace}

\newcommand{\bigclam}{BigCLAM\xspace}
\newcommand{\cesna}{CESNA\xspace}
\newcommand{\cde}{CDE\xspace}
\newcommand{\snmf}{SNMF\xspace}
\newcommand{\snetoc}{SNetOC\xspace}
\newcommand{\epm}{EPM\xspace}

\newcommand{\gtgkm}{G2G/NEO\xspace}
\newcommand{\gtg}{Graph2Gauss\xspace}
\newcommand{\dwkm}{DW/NEO\xspace}
\newcommand{\neokmeans}{NEO-K-Means\xspace}
\newcommand{\relu}{\textnormal{ReLU}}
\newcommand{\GCN}{\operatorname{GCN}}

\newcommand{\MLP}{\operatorname{MLP}}
\newcommand{\bernoulli}{\operatorname{Bernoulli}}

\newcommand{\magcs}{Computer Science\xspace}
\newcommand{\magchem}{Chemistry\xspace}
\newcommand{\magmed}{Medicine\xspace}
\newcommand{\mageng}{Engineering\xspace}

\def\Figref#1{Figure~\ref{#1}}

\def\Secref#1{Section~\ref{#1}}

\def\eqref#1{equation~\ref{#1}}
\def\Eqref#1{Equation~\ref{#1}}

\def\1{\bm{1}}

\def\vmu{{\bm{\mu}}}
\def\vtheta{{\bm{\theta}}}

\def\mA{{\bm{A}}}

\def\mD{{\bm{D}}}

\def\mF{{\bm{F}}}
\def\mG{{\bm{G}}}

\def\mI{{\bm{I}}}

\def\mW{{\bm{W}}}
\def\mX{{\bm{X}}}

\DeclareMathAlphabet{\mathsfit}{\encodingdefault}{\sfdefault}{m}{sl}
\SetMathAlphabet{\mathsfit}{bold}{\encodingdefault}{\sfdefault}{bx}{n}

\def\gL{{\mathcal{L}}}

\def\gS{{\mathcal{S}}}

\newcommand{\E}{\mathbb{E}}

\newcommand{\R}{\mathbb{R}}

\DeclareMathOperator*{\argmin}{arg\,min}

\usepackage{hyperref}
\usepackage{url}

\setcopyright{rightsretained}

\acmConference[DLG'19]{The First International Workshop on Deep Learning for Graphs}{August 2019}{Anchorage, Alaska, USA}
\acmYear{2019}
\copyrightyear{2019}

\begin{document}
\title[]{Overlapping Community Detection\\ with Graph Neural Networks}

\author{Oleksandr Shchur}
\affiliation{%
  \institution{Technical Univeristy of Munich, Germany}
}
\email{shchur@in.tum.de}

\author{Stephan G\"unnemann}
\affiliation{%
  \institution{Technical Univeristy of Munich, Germany}
}
\email{guennemann@in.tum.de}

\begin{abstract}
    Community detection is a fundamental problem in machine learning.
    While deep learning has shown great promise in many graph-related tasks,
    developing neural models for community detection has received surprisingly little attention.
    The few existing approaches focus on detecting disjoint communities, even though communities in real graphs are well known to be overlapping.
    We address this shortcoming and propose a graph neural network (GNN) based model for overlapping community detection.
    Despite its simplicity, our model outperforms the existing baselines by a large margin in the task of community recovery.
    We establish through an extensive experimental evaluation that the proposed model is effective, scalable and robust to hyperparameter settings.
    We also perform an ablation study that confirms that GNN is the key ingredient to the power of the proposed model.
    A reference implementation as well as the new datasets are available under \url{www.kdd.in.tum.de/nocd}.

\end{abstract}

\maketitle
\section{Introduction}
\label{sec:introduction}
Graphs provide a natural way of representing complex real-world systems.
Community detection methods are an essential tool for understanding the structure and behavior of these systems.
Detecting communities allows us to analyze social networks \citep{girvan2002social},
detect fraud \citep{pinheiro2012fraud},
discover functional units of the brain \citep{garcia2018brain}, and predict functions of proteins \citep{song2009protein}.
The problem of community detection has attracted significant attention of the research community and numerous models and algorithms have been proposed \citep{xie2013survey}.

In the recent years, the emerging field of deep learning for graphs has shown great promise in designing more accurate and more scalable algorithms.
While deep learning approaches have achieved unprecedented results in graph-related tasks like link prediction and node classification \cite{cai2018embeddingsurvey},
relatively little attention has been dedicated to their application for unsupervised community detection.
Several methods have been proposed \citep{yang2016deepmodularity, choong2018vgaecd, cavallari2017come},
but they all have a common drawback:
they only focus on the special case of disjoint (non-overlapping) communities.
However, it is well known that communities in real networks are overlapping \cite{yang2014structure}.
Handling overlapping communities is a requirement not yet met by existing deep learning approaches for community detection.

In this paper we address this research gap and propose an end-to-end deep learning model capable of detecting overlapping communities.
To summarize, our main contributions are:
\begin{itemize}
	\item Model:
	We introduce a graph neural network (GNN) based model for overlapping community detection.
	\item Data: We introduce 4 new datasets for overlapping community detection that can act as a benchmark and stimulate future research in this area.
	\item Experiments: We perform a thorough evaluation of our model
	and show its superior performance compared to established methods for overlapping community detection,
	both in terms of speed and accuracy.
	We highlight the importance of the GNN component of our model through an ablation study.
\end{itemize}

\section{Background}
\label{sec:community-detection}

Assume that we are given an undirected unweighted graph $\mG$, represented as a binary adjacency matrix $\mA \in \{0, 1\}^{N \times N}$.
We denote as $N$ the number of nodes $V = \{1, ..., N\}$; and as $M$ the number of edges $E = \{(u, v) \in V \times V : A_{uv} = 1\}$.
Every node might be associated with a $D$-dimensional attribute vector, that can be represented as an attribute matrix $\mX \in \R^{N \times D}$.
The goal of overlapping community detection is to assign nodes into $C$ communities.
Such assignment can be represented as a non-negative community affiliation matrix $\mF \in \R_{\ge 0}^{N \times C}$,
where $F_{uc}$ denotes the strength of node $u$'s membership in community $c$
(with the notable special case of binary assignment $\mF \in \{0, 1\}^{N \times C}$).
Some nodes may be assigned to no communities, while others may belong to multiple.

Even though the notion of "community" seems rather intuitive, there is no universally agreed upon definition of it in the literature.
However, most recent works tend to agree with the statement that a community is a group of nodes
that have higher probability to form edges with each other than with other nodes in the graph \citep{fortunato2016community}.
This way, the problem of community detection can be considered in terms of the probabilistic inference framework.
Once we posit a community-based generative model $p(\mG | \mF)$ for the graph,
detecting communities boils down to inferring the unobserved affiliation matrix $\mF$ given the observed graph $\mG$.

Besides the traditional probabilistic view, one can also view community detection through the lens of representation learning.
The community affiliation matrix $\mF$ can be considered as an embedding of nodes into $\R_{\ge 0}^C$, with the aim of preserving the graph structure.
Given the recent success of representation learning for graphs \citep{cai2018embeddingsurvey}, a question arises:
"Can the advances in deep learning for graphs be used to design better community detection algorithms?".
As we show in \Secref{sec:exp-recovery}, simply combining existing node embedding approaches with overlapping K-means doesn't lead to satisfactory results.
Instead, we propose to combine the probabilistic and representation points of view, and learn the community affiliations in an end-to-end manner using a graph neural network.

\section{The \model model}
\label{sec:model}
Here, we present the Neural Overlapping Community Detection (\model) model.
The core idea of our approach is to combine the power of GNNs with the Bernoulli--Poisson probabilistic model. %

\subsection{Bernoulli--Poisson model}
The Bernoulli--Poisson (BP) model \cite{yang2013bigclam,zhou2015bplink,todeschini2016exchangeable}
is a graph generative model that allows for overlapping communities.
According to the BP model, the graph is generated as follows.
Given the affiliations $\mF \in \R_{\ge 0}^{N \times C}$, adjacency matrix entries $A_{uv}$ are sampled i.i.d. as
\begin{align}
    \label{eq:bigclam}
    A_{uv} \sim \bernoulli(1 - \exp(-\mF_u \mF_v^T))
\end{align}
where $\mF_u$ is the row vector of community affiliations of node $u$ (the $u$'s row of the matrix $\mF$).
Intuitively, the more communities nodes $u$ and $v$ have in common (i.e. the higher the dot product $\mF_u \mF_v^T$ is),
the more likely they are to be connected by an edge.

This model has a number of desirable properties: It can produce various community topologies (e.g. nested, hierarchical),
leads to dense overlaps between communities \cite{yang2014structure} and is computationally efficient (\Secref{sec:model-scalability}).
Existing works propose to perform inference in the BP model using maximum likelihood estimation with coordinate ascent \cite{yang2013bigclam,yang2013cesna}
or Markov chain Monte Carlo \cite{zhou2015bplink,todeschini2016exchangeable}.

\subsection{Model definition}
\label{sec:model-architecture}
Instead of treating the affiliation matrix $\mF$ as a free variable over which optimization is performed,
we generate $\mF$ with a GNN:
\begin{align}
    \label{eq:model}
    \mF := \operatorname{GNN}_{\vtheta} (\mA, \mX)
\end{align}
A ReLU nonlinearity is applied element-wise to the output layer to ensure non-negativity of $\mF$.
See \Secref{sec:evaluation} and Appendix \ref{app:architecture} for details about the GNN architecture.

The negative log-likelihood of the Bernoulli--Poisson model is
\begin{align}
    \label{eq:bp-nll}
    -\log p(\mA | \mF) &= -\hspace{-3mm}\sum_{(u, v) \in E} \log (1 - \exp(-\mF_u \mF_v^T)) + \hspace{-2mm}\sum_{(u, v) \notin E} \mF_u \mF_v^T
\end{align}
Real-world graph are usually extremely sparse,
which means that the second term in \Eqref{eq:bp-nll} will provide a much larger contribution to the loss.
We counteract this by balancing the two terms, which is a standard technique in imbalanced classification \cite{he2008imbalanced}
\begin{align}
    \label{eq:loss}
    \hspace{-1mm}\gL(\mF) &= \hspace{-0.5mm}-\E_{(u, v) \sim P_E} \hspace{-1mm}\left[\log (1 \hspace{-0.5mm}- \hspace{-0.5mm}\exp(-\mF_u \mF_v^T))\right]
    \hspace{-0.5mm}+ \hspace{-0.5mm} \E_{(u, v) \sim P_N} \hspace{-1mm}\left[\mF_u \mF_v^T\right]
\end{align}
where $P_E$ and $P_N$ denote uniform distributions over edges and non-edges respectively.

Instead of directly optimizing the affiliation matrix $\mF$, as done by traditional approaches \cite{yang2013bigclam,yang2013cesna},
we search for neural network parameters $\vtheta^{\star}$ that minimize the (balanced) negative log-likelihood
\begin{align}
    \vtheta^\star = \argmin_{\vtheta} \gL(\operatorname{GNN}_{\vtheta} (\mA, \mX))
\end{align}

Using a GNN for community prediction has several advantages.
First, due to an appropriate inductive bias, the GNN outputs similar community affiliation vectors for neighboring nodes,
which improves the quality of predictions compared to simpler models (\Secref{sec:exp-simple}).
Also, such formulation allows us to seamlessly incorporate the node features into the model.
If the node attributes $\mX$ are not available, we can simply use $\mA$ as node features \cite{kipf2016gcn}.
Finally, with the formulation from \Eqref{eq:model}, it's even possible to predict communities inductively for nodes not seen at training time.

\subsection{Scalability}
\label{sec:model-scalability}
One advantage of the BP model is that it allows to efficiently evaluate the loss $\gL(\mF)$ and its gradients w.r.t. $\mF$.
By using a caching trick \cite{yang2013bigclam}, we can reduce the computational complexity of these operations from $O(N^2)$ to $O(N + M)$.
While this already leads to large speed-ups due to sparsity of real-world networks (typically $M \ll N^2$ ), we can speed it up even further.
Instead of using all entries of $\mA$ when computing the loss (\Eqref{eq:loss}), we sample a mini-batch of $S$ edges and non-edges at each training epoch,
thus approximately computing $\nabla \gL$ in $O(S)$.
In Appendix \ref{app:convergence} we show that this stochastic optimization strategy converges to the same solution as the full-batch approach,
while keeping the computational cost and memory footprint low.

While we subsample the graph to efficiently evaluate the training objective $\gL(\mF)$, we use the full adjacency matrix inside the GNN.
This doesn't limit the scalability of our model:
\model is trained on a graph with 800K+ edges in 3 minutes on a single GPU (see \Secref{sec:exp-recovery}).
It is straightforward to make the GNN component even more scalable by applying the techniques such as \cite{chen2018stochastic,ying2018pinsage}.

\section{Evaluation}
\label{sec:evaluation}

\begin{table*}[t]
	\caption{Recovery of ground-truth communities, measured by NMI (in \%).
	Results for \model are averaged over 50 initializations (see Table 4 for error bars).
	Best result for each row in \textbf{bold}.
	DNF --- did not finish in 12 hours or ran out of memory.}
	\label{tab:recovery}
	\begin{center}
	\vspace{-3mm}
	\resizebox{0.88\textwidth}{!}{%
	\begin{tabular}{lrrrrrrrrrrr}
		Dataset       & \bigclam & \cesna & \epm & \snetoc & \cde  & \snmf & \dwkm & \gtgkm & \modelA & \modelX \\
		\toprule
		Facebook 348  &     26.0 &   29.4 &  6.5 &    24.0   &  24.8   &  13.5 &  31.2 &  17.2 &    34.7 &    \textbf{36.4} \\
		Facebook 414  &     48.3 &   50.3 &  17.5 &   52.0 &  28.7 &  32.5 &  40.9 &   32.3 &    56.3 &    \textbf{59.8} \\
		Facebook 686  &     13.8 &   13.3 &   3.1 &   10.6 &  13.5 &  11.6 &  11.8 &    5.6 &    20.6 &    \textbf{21.0} \\
		Facebook 698  &     45.6 &   39.4 &   9.2 &   44.9 &  31.6 &  28.0 &  40.1 &    2.6 &    \textbf{49.3} &    41.7 \\
		Facebook 1684 &     32.7 &   28.0 &   6.8 &   26.1 &  28.8 &  13.0 &  \textbf{37.2} &    9.9 &    34.7 &    26.1 \\
		Facebook 1912 &     21.4 &   21.2 &   9.8 &   21.4 &  15.5 &  23.4 &  20.8 &   16.0 &    \textbf{36.8} &    35.6 \\
		\magchem       &      0.0 &   23.3 &   DNF &    DNF &   DNF &   2.6 &   1.7 &   22.8 &    22.6 &    \textbf{45.3} \\
		\magcs         &      0.0 &   33.8 &   DNF &    DNF &   DNF &   9.4 &   3.2 &   31.2 &    34.2 &    \textbf{50.2} \\
		\mageng        &      7.9 &   24.3 &   DNF &    DNF &   DNF &  10.1 &   4.7 &   33.4 &    18.4 &    \textbf{39.1} \\
		\magmed        &      0.0 &   14.4 &   DNF &    DNF &   DNF &   4.9 &   5.5 &   28.8 &    27.4 &    \textbf{37.8} \\
	\end{tabular}%
	}
	\vspace{-4mm}
	\end{center}
\end{table*}

\textbf{Datasets. }
We use the following real-world graph datasets in our experiments.
\textbf{Facebook} \citep{mcauley2014circles} is a collection of small (50-800 nodes) ego-networks from the Facebook graph.
Larger graph datasets (10K+ nodes) with reliable ground-truth overlapping community information
and node attributes are not openly available,
which hampers the evaluation and development of new methods. %
For this reason we have collected and preprocessed 4 real-world datasets,
that satisfy these criteria and can act as future benchmarks.
\textbf{\magchem}, \textbf{\magcs}, \textbf{\magmed}, \textbf{\mageng} are co-authorship networks, constructed from the Microsoft Academic Graph \cite{mag}.
Communities correspond to research areas in respective fields,
and node attributes are based on keywords of the papers by each author.
Statistics for all used datasets are provided in Appendix \ref{app:datasets}.

\textbf{Model architecture.}
For all experiments, we use a 2-layer Graph Convolutional Network (GCN) \cite{kipf2016gcn} as the basis for the \model model.
The GCN is defined as
\begin{align}
	\label{eq:gcn}
	\mF := \GCN_{\vtheta}(\mA, \mX) = \relu(\hat{\mA}~\relu(\hat{\mA} \mX \mW^{(1)})\mW^{(2)})
\end{align}
where $\hat{\mA} = \tilde{\mD}^{-1/2} \tilde{\mA} \tilde{\mD}^{-1/2}$ is the normalized adjacency matrix,
$\tilde{\mA} = \mA + \mI_{N}$ is the adjacency matrix with self loops, and $\tilde{D}_{ii} = \sum_{j} \tilde{A}_{ij}$ is the diagonal degree matrix of $\tilde{\mA}$.
We considered other GNN architectures, as well as deeper models, but none of them led to any noticeable improvements.
The two main differences of our model from standard GCN include (1) batch normalization after the first graph convolution layer and
(2) $L_2$ regularization applied to all weight matrices.
We found both of these modifications to lead to substantial gains in performance.
We optimized the architecture and hyperparameters only using the \magcs dataset --- no additional tuning was done for other datasets.
More details about the model configuration and the training procedure are provided in Appendix \ref{app:architecture}.
We denote the model working on node attributes $\mX$ as \modelX, and the model using the adjacency matrix $\mA$ as input as \modelA.
In both cases, the feature matrix is row-normalized.

\textbf{Assigning nodes to communities. }
In order to compare the detected communities to the ground truth,
we first need to convert the predicted continuous community affiliations $\mF$ into binary community assignments.
We assign node $u$ to community $c$ if its affiliation strength $F_{uc}$ is above a fixed threshold $\rho$.
We chose the threshold $\rho=0.5$ like all other hyperparameters --- by picking the value that achieved the best score on the \magcs dataset,
and then using it in further experiments without additional tuning.

\textbf{Metrics. }
We found that popular metrics for quantifying agreement between true and detected communities,
such as Jaccard and $F_1$ scores \cite{yang2013bigclam,yang2013cesna,li2018cde}, can give arbitrarily high scores for completely uninformative community assignments.
See Appendix \ref{app:metrics} for an example and discussion.
Instead we use overlapping normalized mutual information (NMI) \cite{mcdaid2011normalized}, as it is more robust and meaningful.

\subsection{Recovery of ground-truth communities}
\label{sec:exp-recovery}
We evaluate the \model model by checking how well it recovers communities in graphs with known ground-truth communities.

\textbf{Baselines. }
In our selection of baselines, we chose methods that are based on different paradigms for overlapping community detection:
probabilistic inference, non-negative matrix factorization (NMF) and deep learning.
Some methods incorporate the attributes, while other rely solely on the graph structure.

\bigclam \citep{yang2013bigclam}, \epm \cite{zhou2015bplink} and \snetoc \cite{todeschini2016exchangeable} are based on the Bernoulli--Poisson model.
\bigclam learns $\mF$ using coordinate ascent, while \epm and \snetoc perform inference with Markov chain Monte Carlo (MCMC).
\cesna \citep{yang2013cesna} is an extension of \bigclam that additionally models node attributes.
\snmf \citep{wang2011nmf} and \cde \citep{li2018cde} are NMF approaches for overlapping community detection.

We additionally implemented two methods based on neural graph embedding.
First, we compute node embeddings for all the nodes in the given graph using two established approaches --
\deepwalk \cite{perozzi2014deepwalk} and \gtg \cite{bojchevski2018g2g}.
\gtg takes into account both node features and the graph structure, while \deepwalk only uses the structure.
Then, we cluster the nodes using Non-exhaustive Overlapping (NEO) K-Means \cite{whang2015neo} --- which allows to assign them to overlapping communities.
We denote the methods based on \deepwalk and \gtg as \dwkm and \gtgkm respectively.

To ensure a fair comparison, all methods were given the true number of communities $C$.
Other hyperparameters were set to their recommended values.
An overview of all baseline methods, as well as their configurations are provided in Appendix \ref{app:baselines}.

\begin{table*}[t]
	\caption{Comparison of the GNN-based model against simpler baselines.
	Multilayer perceptron (MLP) and Free Variable (FV) models are optimizing the same objective (\Eqref{eq:loss}), but represent the community affiliations $\mF$ differently.}
	\label{tab:simple}
	\begin{center}
	\vspace{-3mm}
	\begin{tabular}{lcccccc}
	{}             &  \multicolumn{2}{c}{Attributes}          &  \multicolumn{2}{c}{Adjacency}         &                   \\
	Dataset        &  GNN               &  MLP               &  GNN               &  MLP               &  Free variable    \\
	\toprule
		Facebook 348  &  \textbf{36.4 $\pm$ 2.0} &  11.7 $\pm$ 2.7 &  34.7 $\pm$ 1.5 &  27.7 $\pm$ 1.6 &  25.7 $\pm$ 1.3 \\
		Facebook 414  &  \textbf{59.8 $\pm$ 1.8} &  22.1 $\pm$ 3.1 &  56.3 $\pm$ 2.4 &  48.2 $\pm$ 1.7 &  49.2 $\pm$ 0.4 \\
		Facebook 686  &  \textbf{21.0 $\pm$ 0.9} &   1.5 $\pm$ 0.7 &  20.6 $\pm$ 1.4 &  19.8 $\pm$ 1.1 &  13.5 $\pm$ 0.9 \\
		Facebook 698  &  41.7 $\pm$ 3.6 &   1.4 $\pm$ 1.3 &  \textbf{49.3 $\pm$ 3.4} &  42.2 $\pm$ 2.7 &  41.5 $\pm$ 1.5 \\
		Facebook 1684 &  26.1 $\pm$ 1.3 &  17.1 $\pm$ 2.0 &  \textbf{34.7 $\pm$ 2.6} &  31.9 $\pm$ 2.2 &  22.3 $\pm$ 1.4 \\
		Facebook 1912 &  35.6 $\pm$ 1.3 &  17.5 $\pm$ 1.9 &  \textbf{36.8 $\pm$ 1.6} &  33.3 $\pm$ 1.4 &  18.3 $\pm$ 1.2 \\
		\magchem      &  45.3 $\pm$ 2.3 &  \textbf{46.6 $\pm$ 2.9} &  22.6 $\pm$ 3.0 &  12.1 $\pm$ 4.0 &   5.2 $\pm$ 2.3 \\
		\magcs        &  \textbf{50.2 $\pm$ 2.0} &  49.2 $\pm$ 2.0 &  34.2 $\pm$ 2.3 &  31.9 $\pm$ 3.8 &  15.1 $\pm$ 2.2 \\
		\mageng       &  39.1 $\pm$ 4.5 &  \textbf{44.5 $\pm$ 3.2} &  18.4 $\pm$ 1.9 &  15.8 $\pm$ 2.1 &   7.6 $\pm$ 2.2 \\
		\magmed       &  \textbf{37.8 $\pm$ 2.8} &  31.8 $\pm$ 2.1 &  27.4 $\pm$ 2.5 &  23.6 $\pm$ 2.1&   9.4 $\pm$ 2.3 \\
	\end{tabular}
	\vspace{-4mm}
	\end{center}
\end{table*}

\textbf{Results: Recovery. }
Table \ref{tab:recovery} shows how well different methods recover the ground-truth communities.
Either \modelX or \modelA achieve the highest score for 9 out of 10 datasets.
We found that the NMI of both methods is strongly correlated with the reconstruction loss (\Eqref{eq:loss}):
\modelA outperforms \modelX in terms of NMI exactly in those cases, when \modelA achieves a lower reconstruction loss.
This means that we can pick the better performing of two methods in a completely unsupervised fashion by only considering the loss values.

\textbf{Results: Hyperparameter sensitivity. }
It's worth noting again that both \model models use the same hyperparameter configuration that was tuned only on the \magcs dataset ($N=22K, M=96.8K, D=7.8K$).
Nevertheless, both models achieve excellent results on datasets with dramatically different characteristics (e.g. Facebook 414 with $N=150, M=1.7K, D=16$).

\textbf{Results: Scalability. }
In addition to displaying excellent recovery results, \model is highly scalable.
\model is trained on the \magmed dataset (63K nodes, 810K edges) using a single GTX1080Ti GPU in 3 minutes,
while only using 750MB of GPU RAM (out of 11GB available).
See Appendix \ref{app:hardware} for more details on hardware.

\epm, \snetoc and \cde don't scale to larger datasets, since they instantiate very large dense matrices during computations.
\snmf and \bigclam, while being the most scalable methods and having lower runtime than \model,
achieved relatively low scores in recovery.
Generating the embeddings with \deepwalk and \gtg can be done very efficiently.
However, overlapping clustering of the embeddings with \neokmeans was the bottleneck, which led to runtimes exceeding several hours for the large datasets.
As the authors of \cesna point out \cite{yang2013cesna}, the method scales to large graphs if the number of attributes $D$ is low.
However, as $D$ increases, which is common for modern datasets, the method scales rather poorly.
This is confirmed by our findings --- on the \magmed dataset, \cesna (parallel version with 18 threads) took 2 hours to converge.

\subsection{Do we really need a graph neural network?}
\label{sec:exp-simple}
Our GNN-based model achieved superior performance in community recovery.
Intuitively, it makes sense to use a GNN for the reasons laid out in \Secref{sec:model-architecture}.
Nevertheless, we should ask whether it's possible achieve comparable results with a simpler model.
To answer this question, we consider the following two baselines.

\textbf{Multilayer perceptron (MLP):} Instead of a GCN (\Eqref{eq:gcn}), we use a simple fully-connected neural network to generate $\mF$.
\begin{equation}
	\mF = \operatorname{MLP}_\vtheta(\mX) = \relu(\relu(\mX \mW^{(1)}) \mW^{(2)})
	\label{eq:mlp}
\end{equation}
This is related to the model proposed by \cite{hu2017deep}.
Same as for the GCN-based model, we optimize the weights of the MLP, $\vtheta = \{\mW^{(1)}, \mW^{(2)}\}$, to minimize the objective \Eqref{eq:loss}.
\begin{equation}
	\min_{\vtheta} \gL(\MLP_{\theta}(\mX))
	\label{eq:mlp-obj}
\end{equation}

\textbf{Free variable (FV): }
As an even simpler baseline, we consider treating $\mF$ as a free variable in optimization and solve
\begin{equation}
	\label{eq:free-var}
	\min_{\mF \ge 0} \gL(\mF)
\end{equation}
We optimize the objective using projected gradient descent with Adam \cite{kingma2014adam},
and update all the entries of $\mF$ at each iteration.
This can be seen as an improved version of the \bigclam model.
Original \bigclam uses the imbalanced objective (\Eqref{eq:bp-nll}) and
optimizes $\mF$ using coordinate ascent with backtracking line search.

\textbf{Setup. }
Both for the MLP and FV models, we tuned the hyperparameters on the Computer Science dataset (just as we did for the GNN model),
and used the same configuration for all datasets.
Details about the configuration for both models are provided in Appendix \ref{app:architecture}.
Like before, we consider the variants of the GNN-based and MLP-based models that use either $\mX$ or $\mA$ as input features.
We compare the NMI scores obtained by the models on all 11 datasets.

\textbf{Results. }
The results for all models are shown in Table \ref{tab:simple}.
The two neural network based models consistently outperform the free variable model.
When node attributes $\mX$ are used, the MLP-based model outperforms the GNN version for \magchem and \mageng datasets, where the node features alone provide a strong signal.
However, MLP achieves extremely low scores for Facebook 686 and Facebook 698 datasets, where the attributes are not as reliable.
On the other hand, when $\mA$ is used as input, the GNN-based model always outperforms MLP.
Combined, these findings confirm our hypothesis that a graph-based neural network architecture is indeed beneficial for the community detection task.

\section{Related work}
\label{sec:related-work}

The problem of community detection in graphs is well-established in the research literature.
However, most of the works study detection of non-overlapping communities \cite{abbe2018sbm,von2007tutorial}.
Algorithms for overlapping community detection can be broadly divided into methods based on non-negative matrix factorization \cite{li2018cde,wang2011nmf,kuang2012nmf},
probabilistic inference \cite{yang2013bigclam,zhou2015bplink,todeschini2016exchangeable,latouche2011osbm}, and heuristics \cite{gleich2012neighborhoods,galbrun2014overlapping,ruan2013codicil,li2015ospectral}.

Deep learning for graphs can be broadly divided into two categories: graph neural networks and node embeddings.
GNNs \citep{kipf2016gcn, hamilton2017graphsage, xu2018jknet}
are specialized neural network architectures that can operate on graph-structured data.
The goal of embedding approaches \citep{perozzi2014deepwalk, kipf2016gae, grover2016node2vec, bojchevski2018g2g}
is to learn vector representations of nodes in a graph that can then be used for downstream tasks.
While embedding approaches work well for detecting disjoint communities \cite{cavallari2017come,tsitsulin2018verse},
they are not well-suited for overlapping community detection, as we showed in our experiments.
This is caused by lack of reliable and scalable approaches for overlapping clustering of vector data.

Several works have proposed deep learning methods for community detection.
\cite{yang2016deepmodularity} and \cite{cao2018incorporating} use neural nets to factorize the modularity matrix,
while \cite{cavallari2017come} jointly learns embeddings for nodes and communities.
However, neither of these methods can handle overlapping communities.
Also related to our model is the approach by \cite{hu2017deep}, where they use a deep belief network to learn community affiliations.
However, their neural network architecture does not use the graph, which we have shown to be crucial in \Secref{sec:exp-simple};
and, just like \epm and \snetoc, relies on MCMC, which heavily limits the scalability of their approach.
Lastly, \cite{chen2017supervisedcd} designed a GNN for \emph{supervised} community detection,
which is a very different setting.

\section{Discussion \& Future work}
We proposed \model --- a graph neural network model for overlapping community detection.
The experimental evaluation confirms that the model is accurate, flexible and scalable.

Besides strong empirical results, our work opens interesting follow-up questions.
We plan to investigate how the two versions of our model (\modelX and \modelA) can be used to quantify the relevance of attributes to the community structure.
Moreover, we plan to assess the inductive performance of \model \cite{hamilton2017graphsage}.

To summarize, the results obtained in this paper provide strong evidence
that deep learning for graphs deserves more attention as a framework for overlapping community detection.

\section*{Acknowledgments}
This research was supported by the German Research Foundation, Emmy Noether grant GU 1409/2-1.

\bibliography{bibliography}


\begin{thebibliography}{45}


\ifx \showCODEN    \undefined \def \showCODEN     #1{\unskip}     \fi
\ifx \showDOI      \undefined \def \showDOI       #1{#1}\fi
\ifx \showISBNx    \undefined \def \showISBNx     #1{\unskip}     \fi
\ifx \showISBNxiii \undefined \def \showISBNxiii  #1{\unskip}     \fi
\ifx \showISSN     \undefined \def \showISSN      #1{\unskip}     \fi
\ifx \showLCCN     \undefined \def \showLCCN      #1{\unskip}     \fi
\ifx \shownote     \undefined \def \shownote      #1{#1}          \fi
\ifx \showarticletitle \undefined \def \showarticletitle #1{#1}   \fi
\ifx \showURL      \undefined \def \showURL       {\relax}        \fi
\providecommand\bibfield[2]{#2}
\providecommand\bibinfo[2]{#2}
\providecommand\natexlab[1]{#1}
\providecommand\showeprint[2][]{arXiv:#2}

\bibitem[\protect\citeauthoryear{??}{mag}{[n. d.]}]%
        {mag}
 \bibinfo{year}{[n. d.]}\natexlab{}.
\newblock \bibinfo{title}{{Microsoft Academic Graph}}.
\newblock \bibinfo{howpublished}{\url{https://kddcup2016.azurewebsites.net/}}.
\newblock


\bibitem[\protect\citeauthoryear{Abadi, Barham, Chen, Chen, Davis, Dean, Devin,
  Ghemawat, Irving, Isard, et~al\mbox{.}}{Abadi et~al\mbox{.}}{2016}]%
        {abadi2016tensorflow}
\bibfield{author}{\bibinfo{person}{Mart{\'\i}n Abadi}, \bibinfo{person}{Paul
  Barham}, \bibinfo{person}{Jianmin Chen}, \bibinfo{person}{Zhifeng Chen},
  \bibinfo{person}{Andy Davis}, \bibinfo{person}{Jeffrey Dean},
  \bibinfo{person}{Matthieu Devin}, \bibinfo{person}{Sanjay Ghemawat},
  \bibinfo{person}{Geoffrey Irving}, \bibinfo{person}{Michael Isard},
  {et~al\mbox{.}}} \bibinfo{year}{2016}\natexlab{}.
\newblock \showarticletitle{Tensorflow: A system for large-scale machine
  learning}. In \bibinfo{booktitle}{\emph{12th $\{$USENIX$\}$ Symposium on
  Operating Systems Design and Implementation ($\{$OSDI$\}$ 16)}}.
\newblock


\bibitem[\protect\citeauthoryear{Abbe}{Abbe}{2018}]%
        {abbe2018sbm}
\bibfield{author}{\bibinfo{person}{Emmanuel Abbe}.}
  \bibinfo{year}{2018}\natexlab{}.
\newblock \showarticletitle{Community Detection and Stochastic Block Models:
  Recent Developments}.
\newblock \bibinfo{journal}{\emph{JMLR}}  \bibinfo{volume}{18}
  (\bibinfo{year}{2018}).
\newblock


\bibitem[\protect\citeauthoryear{Bojchevski and G\"unnemann}{Bojchevski and
  G\"unnemann}{2018}]%
        {bojchevski2018g2g}
\bibfield{author}{\bibinfo{person}{Aleksandar Bojchevski} {and}
  \bibinfo{person}{Stephan G\"unnemann}.} \bibinfo{year}{2018}\natexlab{}.
\newblock \showarticletitle{Deep {Gaussian} Embedding of Graphs: Unsupervised
  Inductive Learning via Ranking}. In \bibinfo{booktitle}{\emph{ICLR}}.
\newblock


\bibitem[\protect\citeauthoryear{Cai, Zheng, and Chang}{Cai
  et~al\mbox{.}}{2018}]%
        {cai2018embeddingsurvey}
\bibfield{author}{\bibinfo{person}{Hongyun Cai}, \bibinfo{person}{Vincent~W
  Zheng}, {and} \bibinfo{person}{Kevin Chang}.}
  \bibinfo{year}{2018}\natexlab{}.
\newblock \showarticletitle{A comprehensive survey of graph embedding:
  problems, techniques and applications}.
\newblock \bibinfo{journal}{\emph{TKDD}} (\bibinfo{year}{2018}).
\newblock


\bibitem[\protect\citeauthoryear{Cao, Jin, Yang, and Dang}{Cao
  et~al\mbox{.}}{2018}]%
        {cao2018incorporating}
\bibfield{author}{\bibinfo{person}{Jinxin Cao}, \bibinfo{person}{Di Jin},
  \bibinfo{person}{Liang Yang}, {and} \bibinfo{person}{Jianwu Dang}.}
  \bibinfo{year}{2018}\natexlab{}.
\newblock \showarticletitle{Incorporating network structure with node contents
  for community detection on large networks using deep learning}.
\newblock \bibinfo{journal}{\emph{Neurocomputing}}  \bibinfo{volume}{297}
  (\bibinfo{year}{2018}).
\newblock


\bibitem[\protect\citeauthoryear{Cavallari, Zheng, Cai, Chang, and
  Cambria}{Cavallari et~al\mbox{.}}{2017}]%
        {cavallari2017come}
\bibfield{author}{\bibinfo{person}{Sandro Cavallari},
  \bibinfo{person}{Vincent~W Zheng}, \bibinfo{person}{Hongyun Cai},
  \bibinfo{person}{Kevin Chen-Chuan Chang}, {and} \bibinfo{person}{Erik
  Cambria}.} \bibinfo{year}{2017}\natexlab{}.
\newblock \showarticletitle{Learning community embedding with community
  detection and node embedding on graphs}. In \bibinfo{booktitle}{\emph{CIKM}}.
\newblock


\bibitem[\protect\citeauthoryear{Chen, Zhu, and Song}{Chen
  et~al\mbox{.}}{2018}]%
        {chen2018stochastic}
\bibfield{author}{\bibinfo{person}{Jianfei Chen}, \bibinfo{person}{Jun Zhu},
  {and} \bibinfo{person}{Le Song}.} \bibinfo{year}{2018}\natexlab{}.
\newblock \showarticletitle{Stochastic Training of Graph Convolutional Networks
  with Variance Reduction.}. In \bibinfo{booktitle}{\emph{ICML}}.
\newblock


\bibitem[\protect\citeauthoryear{Chen, Li, and Bruna}{Chen
  et~al\mbox{.}}{2019}]%
        {chen2017supervisedcd}
\bibfield{author}{\bibinfo{person}{Zhengdao Chen}, \bibinfo{person}{Xiang Li},
  {and} \bibinfo{person}{Joan Bruna}.} \bibinfo{year}{2019}\natexlab{}.
\newblock \showarticletitle{Supervised Community Detection with Hierarchical
  Graph Neural Networks}. In \bibinfo{booktitle}{\emph{ICLR}}.
\newblock


\bibitem[\protect\citeauthoryear{Choong, Liu, and Murata}{Choong
  et~al\mbox{.}}{2018}]%
        {choong2018vgaecd}
\bibfield{author}{\bibinfo{person}{Jun~Jin Choong}, \bibinfo{person}{Xin Liu},
  {and} \bibinfo{person}{Tsuyoshi Murata}.} \bibinfo{year}{2018}\natexlab{}.
\newblock \showarticletitle{Learning community structure with variational
  autoencoder}. In \bibinfo{booktitle}{\emph{ICDM}}.
\newblock


\bibitem[\protect\citeauthoryear{Fortunato and Hric}{Fortunato and
  Hric}{2016}]%
        {fortunato2016community}
\bibfield{author}{\bibinfo{person}{Santo Fortunato} {and}
  \bibinfo{person}{Darko Hric}.} \bibinfo{year}{2016}\natexlab{}.
\newblock \showarticletitle{Community detection in networks: A user guide}.
\newblock \bibinfo{journal}{\emph{Physics Reports}}  \bibinfo{volume}{659}
  (\bibinfo{year}{2016}).
\newblock


\bibitem[\protect\citeauthoryear{Galbrun, Gionis, and Tatti}{Galbrun
  et~al\mbox{.}}{2014}]%
        {galbrun2014overlapping}
\bibfield{author}{\bibinfo{person}{Esther Galbrun}, \bibinfo{person}{Aristides
  Gionis}, {and} \bibinfo{person}{Nikolaj Tatti}.}
  \bibinfo{year}{2014}\natexlab{}.
\newblock \showarticletitle{Overlapping community detection in labeled graphs}.
\newblock \bibinfo{journal}{\emph{Data Mining and Knowledge Discovery}}
  \bibinfo{volume}{28} (\bibinfo{year}{2014}).
\newblock


\bibitem[\protect\citeauthoryear{Garcia, Ashourvan, Muldoon, Vettel, and
  Bassett}{Garcia et~al\mbox{.}}{2018}]%
        {garcia2018brain}
\bibfield{author}{\bibinfo{person}{Javier~O Garcia}, \bibinfo{person}{Arian
  Ashourvan}, \bibinfo{person}{Sarah Muldoon}, \bibinfo{person}{Jean~M Vettel},
  {and} \bibinfo{person}{Danielle~S Bassett}.} \bibinfo{year}{2018}\natexlab{}.
\newblock \showarticletitle{Applications of community detection techniques to
  brain graphs: Algorithmic considerations and implications for neural
  function}.
\newblock \bibinfo{journal}{\emph{Proc. IEEE}}  \bibinfo{volume}{106}
  (\bibinfo{year}{2018}).
\newblock


\bibitem[\protect\citeauthoryear{Girvan and Newman}{Girvan and Newman}{2002}]%
        {girvan2002social}
\bibfield{author}{\bibinfo{person}{Michelle Girvan} {and}
  \bibinfo{person}{Mark~EJ Newman}.} \bibinfo{year}{2002}\natexlab{}.
\newblock \showarticletitle{Community structure in social and biological
  networks}.
\newblock \bibinfo{journal}{\emph{PNAS}}  \bibinfo{volume}{99}
  (\bibinfo{year}{2002}).
\newblock


\bibitem[\protect\citeauthoryear{Gleich and Seshadhri}{Gleich and
  Seshadhri}{2012}]%
        {gleich2012neighborhoods}
\bibfield{author}{\bibinfo{person}{David~F Gleich} {and} \bibinfo{person}{C
  Seshadhri}.} \bibinfo{year}{2012}\natexlab{}.
\newblock \showarticletitle{Vertex neighborhoods, low conductance cuts, and
  good seeds for local community methods}. In \bibinfo{booktitle}{\emph{KDD}}.
\newblock


\bibitem[\protect\citeauthoryear{Grover and Leskovec}{Grover and
  Leskovec}{2016}]%
        {grover2016node2vec}
\bibfield{author}{\bibinfo{person}{Aditya Grover} {and} \bibinfo{person}{Jure
  Leskovec}.} \bibinfo{year}{2016}\natexlab{}.
\newblock \showarticletitle{node2vec: Scalable feature learning for networks}.
  In \bibinfo{booktitle}{\emph{KDD}}.
\newblock


\bibitem[\protect\citeauthoryear{Hamilton, Ying, and Leskovec}{Hamilton
  et~al\mbox{.}}{2017}]%
        {hamilton2017graphsage}
\bibfield{author}{\bibinfo{person}{Will Hamilton}, \bibinfo{person}{Zhitao
  Ying}, {and} \bibinfo{person}{Jure Leskovec}.}
  \bibinfo{year}{2017}\natexlab{}.
\newblock \showarticletitle{Inductive representation learning on large graphs}.
  In \bibinfo{booktitle}{\emph{NIPS}}.
\newblock


\bibitem[\protect\citeauthoryear{He and Garcia}{He and Garcia}{2008}]%
        {he2008imbalanced}
\bibfield{author}{\bibinfo{person}{Haibo He} {and} \bibinfo{person}{Edwardo~A
  Garcia}.} \bibinfo{year}{2008}\natexlab{}.
\newblock \showarticletitle{Learning from imbalanced data}.
\newblock \bibinfo{journal}{\emph{TKDE}} \bibinfo{number}{9}
  (\bibinfo{year}{2008}).
\newblock


\bibitem[\protect\citeauthoryear{Hu, Rai, and Carin}{Hu et~al\mbox{.}}{2017}]%
        {hu2017deep}
\bibfield{author}{\bibinfo{person}{Changwei Hu}, \bibinfo{person}{Piyush Rai},
  {and} \bibinfo{person}{Lawrence Carin}.} \bibinfo{year}{2017}\natexlab{}.
\newblock \showarticletitle{Deep Generative Models for Relational Data with
  Side Information}.
\newblock \bibinfo{journal}{\emph{ICML}}.
\newblock


\bibitem[\protect\citeauthoryear{Kingma and Ba}{Kingma and Ba}{2015}]%
        {kingma2014adam}
\bibfield{author}{\bibinfo{person}{Diederik~P Kingma} {and}
  \bibinfo{person}{Jimmy Ba}.} \bibinfo{year}{2015}\natexlab{}.
\newblock \showarticletitle{Adam: A method for stochastic optimization}.
\newblock \bibinfo{journal}{\emph{ICLR}} (\bibinfo{year}{2015}).
\newblock


\bibitem[\protect\citeauthoryear{Kipf and Welling}{Kipf and Welling}{2016}]%
        {kipf2016gae}
\bibfield{author}{\bibinfo{person}{Thomas~N Kipf} {and} \bibinfo{person}{Max
  Welling}.} \bibinfo{year}{2016}\natexlab{}.
\newblock \showarticletitle{Variational Graph Auto-Encoders}.
\newblock \bibinfo{journal}{\emph{NIPS Workshop on Bayesian Deep Learning}}.
\newblock


\bibitem[\protect\citeauthoryear{Kipf and Welling}{Kipf and Welling}{2017}]%
        {kipf2016gcn}
\bibfield{author}{\bibinfo{person}{Thomas~N Kipf} {and} \bibinfo{person}{Max
  Welling}.} \bibinfo{year}{2017}\natexlab{}.
\newblock \showarticletitle{Semi-supervised classification with graph
  convolutional networks}.
\newblock \bibinfo{journal}{\emph{ICLR}}.
\newblock


\bibitem[\protect\citeauthoryear{Kuang, Ding, and Park}{Kuang
  et~al\mbox{.}}{2012}]%
        {kuang2012nmf}
\bibfield{author}{\bibinfo{person}{Da Kuang}, \bibinfo{person}{Chris Ding},
  {and} \bibinfo{person}{Haesun Park}.} \bibinfo{year}{2012}\natexlab{}.
\newblock \showarticletitle{Symmetric nonnegative matrix factorization for
  graph clustering}. In \bibinfo{booktitle}{\emph{SDM}}.
\newblock


\bibitem[\protect\citeauthoryear{Latouche, Birmel{\'e}, Ambroise,
  et~al\mbox{.}}{Latouche et~al\mbox{.}}{2011}]%
        {latouche2011osbm}
\bibfield{author}{\bibinfo{person}{Pierre Latouche}, \bibinfo{person}{Etienne
  Birmel{\'e}}, \bibinfo{person}{Christophe Ambroise}, {et~al\mbox{.}}}
  \bibinfo{year}{2011}\natexlab{}.
\newblock \showarticletitle{Overlapping stochastic block models with
  application to the {French} political blogosphere}.
\newblock \bibinfo{journal}{\emph{The Annals of Applied Statistics}}
  \bibinfo{volume}{5} (\bibinfo{year}{2011}).
\newblock


\bibitem[\protect\citeauthoryear{Li, He, Bindel, and Hopcroft}{Li
  et~al\mbox{.}}{2015}]%
        {li2015ospectral}
\bibfield{author}{\bibinfo{person}{Yixuan Li}, \bibinfo{person}{Kun He},
  \bibinfo{person}{David Bindel}, {and} \bibinfo{person}{John~E Hopcroft}.}
  \bibinfo{year}{2015}\natexlab{}.
\newblock \showarticletitle{Uncovering the small community structure in large
  networks: A local spectral approach}. In \bibinfo{booktitle}{\emph{WWW}}.
\newblock


\bibitem[\protect\citeauthoryear{Li, Sha, Huang, and Zhang}{Li
  et~al\mbox{.}}{2018}]%
        {li2018cde}
\bibfield{author}{\bibinfo{person}{Ye Li}, \bibinfo{person}{Chaofeng Sha},
  \bibinfo{person}{Xin Huang}, {and} \bibinfo{person}{Yanchun Zhang}.}
  \bibinfo{year}{2018}\natexlab{}.
\newblock \showarticletitle{Community Detection in Attributed Graphs: An
  Embedding Approach}. In \bibinfo{booktitle}{\emph{AAAI}}.
\newblock


\bibitem[\protect\citeauthoryear{Mcauley and Leskovec}{Mcauley and
  Leskovec}{2014}]%
        {mcauley2014circles}
\bibfield{author}{\bibinfo{person}{Julian Mcauley} {and} \bibinfo{person}{Jure
  Leskovec}.} \bibinfo{year}{2014}\natexlab{}.
\newblock \showarticletitle{Discovering social circles in ego networks}.
\newblock \bibinfo{journal}{\emph{TKDD}}  \bibinfo{volume}{8}
  (\bibinfo{year}{2014}).
\newblock


\bibitem[\protect\citeauthoryear{McDaid, Greene, and Hurley}{McDaid
  et~al\mbox{.}}{2011}]%
        {mcdaid2011normalized}
\bibfield{author}{\bibinfo{person}{Aaron~F McDaid}, \bibinfo{person}{Derek
  Greene}, {and} \bibinfo{person}{Neil Hurley}.}
  \bibinfo{year}{2011}\natexlab{}.
\newblock \showarticletitle{Normalized mutual information to evaluate
  overlapping community finding algorithms}.
\newblock \bibinfo{journal}{\emph{arXiv:1110.2515}} (\bibinfo{year}{2011}).
\newblock


\bibitem[\protect\citeauthoryear{Perozzi, Al-Rfou, and Skiena}{Perozzi
  et~al\mbox{.}}{2014}]%
        {perozzi2014deepwalk}
\bibfield{author}{\bibinfo{person}{Bryan Perozzi}, \bibinfo{person}{Rami
  Al-Rfou}, {and} \bibinfo{person}{Steven Skiena}.}
  \bibinfo{year}{2014}\natexlab{}.
\newblock \showarticletitle{Deepwalk: Online learning of social
  representations}. In \bibinfo{booktitle}{\emph{KDD}}.
\newblock


\bibitem[\protect\citeauthoryear{Pinheiro}{Pinheiro}{2012}]%
        {pinheiro2012fraud}
\bibfield{author}{\bibinfo{person}{Carlos Andr{\'e}~Reis Pinheiro}.}
  \bibinfo{year}{2012}\natexlab{}.
\newblock \showarticletitle{Community detection to identify fraud events in
  telecommunications networks}.
\newblock \bibinfo{journal}{\emph{SAS SUGI proceedings: customer intelligence}}
  (\bibinfo{year}{2012}).
\newblock


\bibitem[\protect\citeauthoryear{Ruan, Fuhry, and Parthasarathy}{Ruan
  et~al\mbox{.}}{2013}]%
        {ruan2013codicil}
\bibfield{author}{\bibinfo{person}{Yiye Ruan}, \bibinfo{person}{David Fuhry},
  {and} \bibinfo{person}{Srinivasan Parthasarathy}.}
  \bibinfo{year}{2013}\natexlab{}.
\newblock \showarticletitle{Efficient community detection in large networks
  using content and links}. In \bibinfo{booktitle}{\emph{WWW}}.
\newblock


\bibitem[\protect\citeauthoryear{Song and Singh}{Song and Singh}{2009}]%
        {song2009protein}
\bibfield{author}{\bibinfo{person}{Jimin Song} {and} \bibinfo{person}{Mona
  Singh}.} \bibinfo{year}{2009}\natexlab{}.
\newblock \showarticletitle{How and when should interactome-derived clusters be
  used to predict functional modules and protein function?}
\newblock \bibinfo{journal}{\emph{Bioinformatics}}  \bibinfo{volume}{25}
  (\bibinfo{year}{2009}).
\newblock


\bibitem[\protect\citeauthoryear{Todeschini, Miscouridou, and Caron}{Todeschini
  et~al\mbox{.}}{2016}]%
        {todeschini2016exchangeable}
\bibfield{author}{\bibinfo{person}{Adrien Todeschini}, \bibinfo{person}{Xenia
  Miscouridou}, {and} \bibinfo{person}{Fran{\c{c}}ois Caron}.}
  \bibinfo{year}{2016}\natexlab{}.
\newblock \showarticletitle{Exchangeable random measures for sparse and modular
  graphs with overlapping communities}.
\newblock \bibinfo{journal}{\emph{arXiv:1602.02114}} (\bibinfo{year}{2016}).
\newblock


\bibitem[\protect\citeauthoryear{Tsitsulin, Mottin, Karras, and
  M{\"u}ller}{Tsitsulin et~al\mbox{.}}{2018}]%
        {tsitsulin2018verse}
\bibfield{author}{\bibinfo{person}{Anton Tsitsulin}, \bibinfo{person}{Davide
  Mottin}, \bibinfo{person}{Panagiotis Karras}, {and} \bibinfo{person}{Emmanuel
  M{\"u}ller}.} \bibinfo{year}{2018}\natexlab{}.
\newblock \showarticletitle{{VERSE}: Versatile Graph Embeddings from Similarity
  Measures}. In \bibinfo{booktitle}{\emph{WWW}}.
\newblock


\bibitem[\protect\citeauthoryear{Von~Luxburg}{Von~Luxburg}{2007}]%
        {von2007tutorial}
\bibfield{author}{\bibinfo{person}{Ulrike Von~Luxburg}.}
  \bibinfo{year}{2007}\natexlab{}.
\newblock \showarticletitle{A tutorial on spectral clustering}.
\newblock \bibinfo{journal}{\emph{Statistics and computing}}
  \bibinfo{volume}{17} (\bibinfo{year}{2007}).
\newblock


\bibitem[\protect\citeauthoryear{Wang, Li, Wang, Zhu, and Ding}{Wang
  et~al\mbox{.}}{2011}]%
        {wang2011nmf}
\bibfield{author}{\bibinfo{person}{Fei Wang}, \bibinfo{person}{Tao Li},
  \bibinfo{person}{Xin Wang}, \bibinfo{person}{Shenghuo Zhu}, {and}
  \bibinfo{person}{Chris Ding}.} \bibinfo{year}{2011}\natexlab{}.
\newblock \showarticletitle{Community discovery using nonnegative matrix
  factorization}.
\newblock \bibinfo{journal}{\emph{Data Mining and Knowledge Discovery}}
  \bibinfo{volume}{22} (\bibinfo{year}{2011}).
\newblock


\bibitem[\protect\citeauthoryear{Whang, Dhillon, and Gleich}{Whang
  et~al\mbox{.}}{2015}]%
        {whang2015neo}
\bibfield{author}{\bibinfo{person}{Joyce~Jiyoung Whang},
  \bibinfo{person}{Inderjit~S Dhillon}, {and} \bibinfo{person}{David~F
  Gleich}.} \bibinfo{year}{2015}\natexlab{}.
\newblock \showarticletitle{Non-exhaustive, Overlapping k-means}. In
  \bibinfo{booktitle}{\emph{SDM}}.
\newblock


\bibitem[\protect\citeauthoryear{Xie, Kelley, and Szymanski}{Xie
  et~al\mbox{.}}{2013}]%
        {xie2013survey}
\bibfield{author}{\bibinfo{person}{Jierui Xie}, \bibinfo{person}{Stephen
  Kelley}, {and} \bibinfo{person}{Boleslaw~K Szymanski}.}
  \bibinfo{year}{2013}\natexlab{}.
\newblock \showarticletitle{Overlapping community detection in networks: The
  state-of-the-art and comparative study}.
\newblock \bibinfo{journal}{\emph{CSUR}}  \bibinfo{volume}{45}
  (\bibinfo{year}{2013}).
\newblock


\bibitem[\protect\citeauthoryear{Xu, Li, Tian, Sonobe, Kawarabayashi, and
  Jegelka}{Xu et~al\mbox{.}}{2018}]%
        {xu2018jknet}
\bibfield{author}{\bibinfo{person}{Keyulu Xu}, \bibinfo{person}{Chengtao Li},
  \bibinfo{person}{Yonglong Tian}, \bibinfo{person}{Tomohiro Sonobe},
  \bibinfo{person}{Ken-ichi Kawarabayashi}, {and} \bibinfo{person}{Stefanie
  Jegelka}.} \bibinfo{year}{2018}\natexlab{}.
\newblock \showarticletitle{Representation learning on graphs with jumping
  knowledge networks}.
\newblock \bibinfo{journal}{\emph{ICML}} (\bibinfo{year}{2018}).
\newblock


\bibitem[\protect\citeauthoryear{Yang and Leskovec}{Yang and Leskovec}{2013}]%
        {yang2013bigclam}
\bibfield{author}{\bibinfo{person}{Jaewon Yang} {and} \bibinfo{person}{Jure
  Leskovec}.} \bibinfo{year}{2013}\natexlab{}.
\newblock \showarticletitle{Overlapping community detection at scale: a
  nonnegative matrix factorization approach}. In
  \bibinfo{booktitle}{\emph{WSDM}}.
\newblock


\bibitem[\protect\citeauthoryear{Yang and Leskovec}{Yang and Leskovec}{2014}]%
        {yang2014structure}
\bibfield{author}{\bibinfo{person}{Jaewon Yang} {and} \bibinfo{person}{Jure
  Leskovec}.} \bibinfo{year}{2014}\natexlab{}.
\newblock \showarticletitle{Structure and Overlaps of Ground-Truth Communities
  in Networks}.
\newblock \bibinfo{journal}{\emph{ACM TIST}}  \bibinfo{volume}{5}
  (\bibinfo{year}{2014}).
\newblock


\bibitem[\protect\citeauthoryear{Yang, McAuley, and Leskovec}{Yang
  et~al\mbox{.}}{2013}]%
        {yang2013cesna}
\bibfield{author}{\bibinfo{person}{Jaewon Yang}, \bibinfo{person}{Julian
  McAuley}, {and} \bibinfo{person}{Jure Leskovec}.}
  \bibinfo{year}{2013}\natexlab{}.
\newblock \showarticletitle{Community detection in networks with node
  attributes}. In \bibinfo{booktitle}{\emph{ICDM}}.
\newblock


\bibitem[\protect\citeauthoryear{Yang, Cao, He, Wang, Wang, and Zhang}{Yang
  et~al\mbox{.}}{2016}]%
        {yang2016deepmodularity}
\bibfield{author}{\bibinfo{person}{Liang Yang}, \bibinfo{person}{Xiaochun Cao},
  \bibinfo{person}{Dongxiao He}, \bibinfo{person}{Chuan Wang},
  \bibinfo{person}{Xiao Wang}, {and} \bibinfo{person}{Weixiong Zhang}.}
  \bibinfo{year}{2016}\natexlab{}.
\newblock \showarticletitle{Modularity Based Community Detection with Deep
  Learning.}. In \bibinfo{booktitle}{\emph{IJCAI}}.
\newblock


\bibitem[\protect\citeauthoryear{Ying, He, Chen, Eksombatchai, Hamilton, and
  Leskovec}{Ying et~al\mbox{.}}{2018}]%
        {ying2018pinsage}
\bibfield{author}{\bibinfo{person}{Rex Ying}, \bibinfo{person}{Ruining He},
  \bibinfo{person}{Kaifeng Chen}, \bibinfo{person}{Pong Eksombatchai},
  \bibinfo{person}{William~L Hamilton}, {and} \bibinfo{person}{Jure Leskovec}.}
  \bibinfo{year}{2018}\natexlab{}.
\newblock \showarticletitle{Graph Convolutional Neural Networks for Web-Scale
  Recommender Systems}.
\newblock


\bibitem[\protect\citeauthoryear{Zhou}{Zhou}{2015}]%
        {zhou2015bplink}
\bibfield{author}{\bibinfo{person}{Mingyuan Zhou}.}
  \bibinfo{year}{2015}\natexlab{}.
\newblock \showarticletitle{Infinite edge partition models for overlapping
  community detection and link prediction}. In
  \bibinfo{booktitle}{\emph{AISTATS}}.
\newblock


\end{thebibliography}
\bibliographystyle{ACM-Reference-Format}

\clearpage

\appendix
\section{Datasets}
\label{app:datasets}
\begin{table}[h]
	\caption{Dataset statistics. $K$ stands for 1000.}
	\label{tab:datasets}
	\begin{center}
	\scalebox{0.9}{
		\begin{tabular}{llrrrr}
		\textbf{Dataset}  &  \textbf{Network type} &      $N$ &       $M$ &      $D$ &   $C$ \\
		\midrule
		Facebook 348      & Social            & $224$   &   $3.2K$  &     $21$ &  14 \\
		Facebook 414      & Social            & $150$   &   $1.7K$  &     $16$ &  7 \\
		Facebook 686      & Social            & $168$   &   $1.6K$  &     $9$ &   14 \\
		Facebook 698      & Social            & $61$    &   $270$   &     $6$ &  13 \\
		Facebook 1684     & Social            & $786$   &   $14.0K$ &     $15$ &  17 \\
		Facebook 1912     & Social            & $747$  &   $30.0K$ &     $29$ &  46 \\
		\magcs            & Co-authorship     & $22.0K$ &   $96.8K$ &   $7.8K$ &  18 \\
		\magchem          & Co-authorship     & $35.4K$ &  $157.4K$ &   $4.9K$ &  14 \\
		\magmed           & Co-authorship     & $63.3K$ &  $810.3K$ &   $5.5K$ &  17 \\
		\mageng           & Co-authorship     & $14.9K$ &   $49.3K$ &   $4.8K$ &  16 \\
	\end{tabular}
}
	\end{center}
\end{table}

\section{Model configuration}
\label{app:architecture}
\subsection{Architecture}
We picked the hyperparameters and chose the model architecture for all 3 models by only considering their performance (NMI) on the \magcs dataset.
No additional tuning for other datasets has been done.

\textbf{GNN-based model. (\Eqref{eq:gcn})}
We use a 2-layer graph convolutional neural network, with hidden size of 128,
and the output (second) layer of size $C$ (number of communities to detect).
We apply batch normalization after the first graph convolution layer.
Dropout with 50\% keep probability is applied before every layer.
We add weight decay to both weight matrices with regularization strength $\lambda = 10^{-2}$.
The feature matrix $\mX$ (or $\mA$, in case we are working without attributes) is normalized such that every row has unit $L_2$-norm.

We also experimented with the Jumping Knowledge Network \cite{xu2018jknet} and GraphSAGE \cite{hamilton2017graphsage} architectures,
but they led to lower NMI scores on the \magcs dataset.

\textbf{MLP-based model. (\Eqref{eq:mlp})}
We found the MLP model to perform best with the same configuration as described above for the GCN model
(i.e. same regularization strength, hidden size, dropout and batch norm).

\textbf{Free variable model (\Eqref{eq:free-var})}
We considered two initialization strategies for the free variable model:
(1) Locally minimal neighborhoods \cite{gleich2012neighborhoods} --- the strategy used by the \bigclam and \cesna models and
(2) initializing $\mF$ to the output of an untrained GCN.
We found strategy (1) to consistently provide better results.

\subsection{Training}
\textbf{GNN- and MLP-based models. }
We train both models using Adam optimizer \cite{kingma2014adam} with default parameters.
The learning rate is set to $10^{-3}$.
We use the following early stopping strategy:
Every 50 epochs we compute the full training loss (\Eqref{eq:loss}).
We stop optimization if there was no improvement in the loss for the last $10 \times 50 = 500$ iterations, or after 5000 epochs, whichever happens first.

\textbf{Free variable model. }
We use Adam optimizer with learning rate $5 \cdot 10^{-2}$.
After every gradient step, we project the $\mF$ matrix to ensure that it stays non-negative: $F_{uc} = \max \{0, F_{uc}\}$.
We use the same early stopping strategy as for the GNN and MLP models.

\section{Baselines}
\label{app:baselines}

\begin{table}[h]
	\caption{Overview of the baselines. See text for the discussion of scalability of \cesna.}
	\label{tab:baselines}
	\begin{center}
		\begin{tabular}{llcc}
			Method & Model type & Attributed & Scalable\\
			\toprule
			\bigclam \cite{yang2013bigclam} & Probabilistic &  &{\large \checkmark} \\
			\cesna \cite{yang2013cesna} & Probabilistic & {\large \checkmark} & ${\large \checkmark}^{*}$\\
			\snetoc \cite{todeschini2016exchangeable} & Probabilistic & \\
			\epm \cite{zhou2015bplink} & Probabilistic & \\
			\cde \cite{li2018cde} & NMF & {\large \checkmark}\\
			\snmf \cite{wang2011nmf} & NMF & & {\large \checkmark}\\
			\dwkm \cite{perozzi2014deepwalk,whang2015neo} & Deep learning & & \\
			\gtgkm \cite{bojchevski2018g2g,whang2015neo} & Deep learning & {\large \checkmark} & \\
			\midrule \model & \begin{tabular}[x]{@{}l@{}}Deep learning + \\probabilistic\end{tabular} & {\large \checkmark} & {\large \checkmark}\\
		\end{tabular}
	\end{center}
\end{table}

\begin{itemize}
\item We used the reference C++ implementations of \bigclam and \cesna, that were provided by the authors (\url{https://github.com/snap-stanford/snap}).
Models were used with the default parameter settings for step size, backtracking line search constants, and balancing terms.
Since \cesna can only handle binary attributes, we binarize the original attributes (set the nonzero entries to 1) if they have a different type.

\item We implemented \snmf ourselves using Python.
The $\mF$ matrix is initialized by sampling from the $\operatorname{Uniform}[0, 1]$ distribution.
We run optimization until the improvement in the reconstruction loss goes below $10^{-4}$ per iteration, or for 300 epochs, whichever happens first.
The results for \snmf are averaged over 50 random initializations.

\item We use the Matlab implementation of \cde provided by the authors.
We set the hyperparameters to $\alpha = 1$, $\beta = 2$, $\kappa = 5$, as recommended in the paper, and run optimization for 20 iterations.

\item For \snetoc and \epm we use the Matlab implementations provided by the authors with the default hyperparameter settings.
The implementation of \epm provides to options: EPM and HEPM. We found EPM to produce better NMI scores, so we used it for all the experiments.

\item We use the TensorFlow implementation of \gtg provided by the authors.
We set the dimension of the embeddings to 128, and only use the $\vmu$ matrix as embeddings.

\item We implemented \deepwalk ourselves: We sample 10 random walks of length 80 from each node, and use the Word2Vec implementation from Gensim (\url{https://radimrehurek.com/gensim/}) to generate the embeddings.The dimension of embeddings is set to 128.

\item For \neokmeans, we use the Matlab code provided by the authors. We let the parameters $\alpha$ and $\beta$ be selected automatically using the built-in procedure.

\end{itemize}

\section{Hardware and software}
\label{app:hardware}
The experiments were performed on a computer running Ubuntu 16.04LTS with 2x Intel(R) Xeon(R) CPU E5-2630 v4 @ 2.20GHz CPUs, 256GB of RAM and 4x GTX1080Ti GPUs.
Note that training and inference were done using only a single GPU at a time for all models.
The \model model was implemented using Tensorflow v1.1.2 \cite{abadi2016tensorflow}

\section{Convergence of the stochastic sampling procedure}
\label{app:convergence}
Instead of using all pairs $u, v \in V$ when computing the gradients $\nabla_{\vtheta} \gL$ at every iteration,
we sample $S$ edges and $S$ non-edges uniformly at random.
We perform the following experiment to ensure that our training procedure
converges to the same result, as when using the full objective.

\textbf{Experimental setup. }
We train the model on the \magcs dataset and compare the full-batch optimization procedure with stochastic gradient descent for different choices of the batch size $S$.
Starting from the same initialization, we measure the full loss (\Eqref{eq:loss}) over the iterations.

\begin{figure}[h]
	\begin{center}
		\includegraphics[width=\columnwidth]{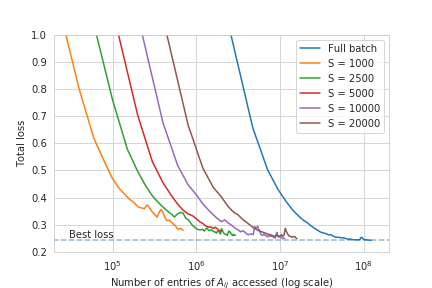}
	\end{center}
\caption{Convergence of the stochastic sampling procedure.}
\label{fig:stochastic}
\end{figure}

\textbf{Results.}
\Figref{fig:stochastic} shows training curves for different batch sizes $S \in \{1000, 2500, 5000, 10000, 20000\}$, as well as for full-batch training.
The horizontal axis of the plot displays the number of entries of adjacency matrix accessed.
One iteration of stochastic training accesses $2S$ entries $A_{ij}$, and one iteration of full-batch accesses $2N + 2M$ entries,
since we are using the caching trick from \cite{yang2013bigclam}.
As we see, the stochastic training procedure is stable:
For batch sizes $S=10K$ and $S=20K$, the loss converges very closely to the value achieved by full-batch training.

\section{Quantifying agreement between overlapping communities}
\label{app:metrics}
A popular choice for quantifying the agreement between true and predicted overlapping communities is the symmetric agreement score \cite{yang2013bigclam,yang2013cesna,li2018cde}.
Given the ground-truth communities $\gS^{*} = \{S_i^*\}_i$ and the predicted communities $\gS = \{S_j\}_j$, the symmetric score is defined as
\begin{align}
    \label{eq:sym-score}
    \frac{1}{2 |\gS^*|} \sum_{S_i^* \in \gS^*} \max_{S_j \in \gS} \delta(S_i^*, S_j) +
    \frac{1}{2 |\gS|}   \sum_{S_j \in \gS} \max_{S_i^* \in \gS^*} \delta(S_i^*, S_j)
\end{align}
where $\delta(S_i^*, S_j)$ is a similarity measure between sets, such as $F_1$-score or Jaccard similarity.

We discovered that these frequently used measures can assign arbitrarily high scores to completely uninformative community assignments,
as you can in see in the following simple example.
Let the ground truth communities be $S_1^* = \{v_1, ..., v_K\}$ and $S_2^* = \{v_{N - K}, ..., v_N\}$ ($K$ nodes in each community),
and let the algorithm assign all the nodes to a single community $S_1 = V = \{v_1, ..., v_N\}$.
While this predicted community assignment is completely uninformative, it will achieve symmetric $F_1$-score of $\frac{2K}{N + K}$ and symmetric Jaccard similarity of $\frac{K}{N}$
(e.g., if $K = 600$ and $N =1000$, the scores will be $75\%$ and $60\%$ respectively).
These high numbers might give a false impression that the algorithm has learned something useful, while that clearly isn't the case.
As an alternative, we suggest using overlapping normalized mutual information (NMI), as defined in \cite{mcdaid2011normalized}.
NMI correctly handles the degenerate cases, like the one above, and assigns them the score of 0.

\end{document}